\documentclass[letterpaper]{article} % DO NOT CHANGE THIS
\usepackage{aaai24}  % DO NOT CHANGE THIS
\usepackage{times}  % DO NOT CHANGE THIS
\usepackage{helvet}  % DO NOT CHANGE THIS
\usepackage{courier}  % DO NOT CHANGE THIS
\usepackage[hyphens]{url}  % DO NOT CHANGE THIS
\usepackage{graphicx} % DO NOT CHANGE THIS
\usepackage{natbib}  % DO NOT CHANGE THIS AND DO NOT ADD ANY OPTIONS TO IT
\usepackage{caption} % DO NOT CHANGE THIS AND DO NOT ADD ANY OPTIONS TO IT
\usepackage{algorithm}
\usepackage{algorithmic}
\usepackage{booktabs}
\usepackage{amssymb}
\usepackage{amsmath}
\newcommand{\model}{SRFormer\ }
 \usepackage{color} 
 \usepackage{dsfont}
\usepackage{multirow}
\usepackage{subcaption}
\usepackage{cleveref}  
\usepackage{colortbl}
\definecolor{mygray}{gray}{.9}
\usepackage{newfloat}
\usepackage{listings}
\DeclareCaptionStyle{ruled}{labelfont=normalfont,labelsep=colon,strut=off} % DO NOT CHANGE THIS
\floatstyle{ruled}
\newfloat{listing}{tb}{lst}{}
\floatname{listing}{Listing}
\newcommand\correspondingauthor{\thanks{Corresponding author.}}

\title{SRFormer: Text Detection Transformer with Incorporated Segmentation and Regression}
\author{
Qingwen Bu\textsuperscript{\rm 1,2}, 
Sungrae Park\textsuperscript{\rm 3}\correspondingauthor, % \textsuperscript{$\dagger$}, 
Minsoo Khang\textsuperscript{\rm 3}, 
Yichuan Cheng\textsuperscript{\rm 4}
}

\affiliations{
\textsuperscript{\rm 1}Shanghai Jiao Tong University, 
\textsuperscript{\rm 2}Shanghai AI Laboratory, \textsuperscript{\rm 3}Upstage AI, \textsuperscript{\rm 4}City University of Hong Kong 

{qwbu01@sjtu.edu.cn, \{sungrae.park, mkhang\}@upstage.ai, yichuan.cheng2000@gmail.com}
}

\usepackage{bibentry}
\begin{document}

\maketitle
\begin{abstract}
Existing techniques for text detection can be broadly classified into two primary groups: segmentation-based and regression-based methods. 
Segmentation models offer enhanced robustness to font variations but require intricate post-processing, leading to high computational overhead. Regression-based methods undertake instance-aware prediction but face limitations in robustness and data efficiency due to their reliance on high-level representations.
In our academic pursuit, we propose SRFormer, a unified DETR-based model with amalgamated \textbf{S}egmentation and \textbf{R}egression, aiming at the synergistic harnessing of the inherent robustness in segmentation representations, along with the straightforward post-processing of instance-level regression. 
Our empirical analysis indicates that favorable segmentation predictions can be obtained at the initial decoder layers. In light of this, we constrain the incorporation of segmentation branches to the first few decoder layers and employ progressive regression refinement in subsequent layers, achieving performance gains while minimizing computational load from the mask.
Furthermore, we propose a Mask-informed Query Enhancement module. We take the segmentation result as a natural soft-ROI to pool and extract robust pixel representations, which are then employed to enhance and diversify instance queries.
Extensive experimentation across multiple benchmarks has yielded compelling findings, highlighting our method's exceptional robustness, superior training and data efficiency, as well as its state-of-the-art performance. Our code is available at \textit{https://github.com/retsuh-bqw/SRFormer-Text-Det}.
\end{abstract}

\label{Sec:intro}
\section{1. Introduction}
Scene text detection and recognition have made many strides in recent years, garnering increasing attention within both the research community and industries, thanks to their wide range of practical applications, such as autonomous driving and document intelligence. 
Despite being a thoroughly investigated area, text detection remains a challenging endeavor within the realm of existing methodologies, particularly when confronted with complex scenarios involving overlapping, irregularly shaped, and stylized text instances.

Previous work on detecting texts can be roughly divided into two streams: regression- and segmentation-based methods. Regression-based methods offer notable advantages, including computational efficiency and adaptability to texts of varying sizes, making them suitable for real-time applications and the detection of both small and large text instances. Additionally, their end-to-end learning approach simplifies the pipeline, enabling post-processing with geometric calculations. However, these methods may exhibit slightly lower localization precision compared to segmentation-based approaches, particularly in the context of irregular or curved text instances~\cite{zhao2019object}. They may also struggle when text contrasts poorly with the surrounding background, making them vulnerable in complex environments.

Segmentation-based models also have their own advantages and limitations. While they can provide pixel-level localization and are more robust in addressing variations in text appearance, such as diverse font styles, sizes, and orientations, they require intricate post-processing to extract complete text instances from the binary masks, involving further algorithmic intervention, which is not amenable to GPU parallel processing~\cite{gu2022review}. This impedes their ability to achieve stable and fast detection.

Can we harness the strength of both regression- and segmentation-based methods, while mitigating their drawbacks by combining these two methods into one unified model? DEtection TRansformers~(DETR), a recent popular method in object detection, present a suitable framework for the integration of these two representations~\cite{li2023mask}. While DETR variants have demonstrated notable success~\cite{zhang2022text,ye2023deepsolo,ye2023dptext}, there remains discernible scope for further enhancement of the performance across various text detection benchmarks. Furthermore, most DETR models adhere to the regression-based paradigm, thereby necessitating prolonged training iterations and substantial datasets to attain optimal performance.

To address the aforementioned issues, we propose SRFormer, a new DETR-based model with separated decoder chunks: the segmentation chunk bootstraps models to learn more robust pixel-level representations, helps the model better separate between text and non-text regions, and provides positional prior for finer-grained regression; the regression chunk directs the queries to capture high-level semantic features and provides further refinement of localization results with minimal post-processing.
Rather than utilizing the segmentation mask directly as the ultimate prediction output, which necessitates accurate prediction and complex post-processing procedures, we introduce a Mask-informed Query Enhancement module that leverages masks as inherent indicators of Regions of Interest (ROI) to extract distinctive features in localized regions, further enhancing and diversifying queries for improved optimization. Utilizing the proposed module alongside supervision signals for both segmentation and regression empowers our model to harness the distinct advantages of each component while alleviating their inherent constraints. 
Our contributions are three folds:

\begin{itemize}
    \item We incorporate regression and segmentation into a unified DETR model, which creates new state-of-the-art performance across several scene text detection benchmarks by leveraging the distinct characteristics from both sides.
    \item Through the strategic incorporation of the segmentation map solely within the initial layers of the decoder, the model gains the capacity to acquire robust pixel-level features and mitigates the need for intricate post-processing steps to ensure a stable and fast inference.
    \item In comparison to regression-based approaches, our proposed method exhibits superior performance in terms of training efficiency, data utilization, as well as improved robustness across diverse data domains.
\end{itemize}

\begin{figure*}[t]
    \centering
    \includegraphics[width=\linewidth]{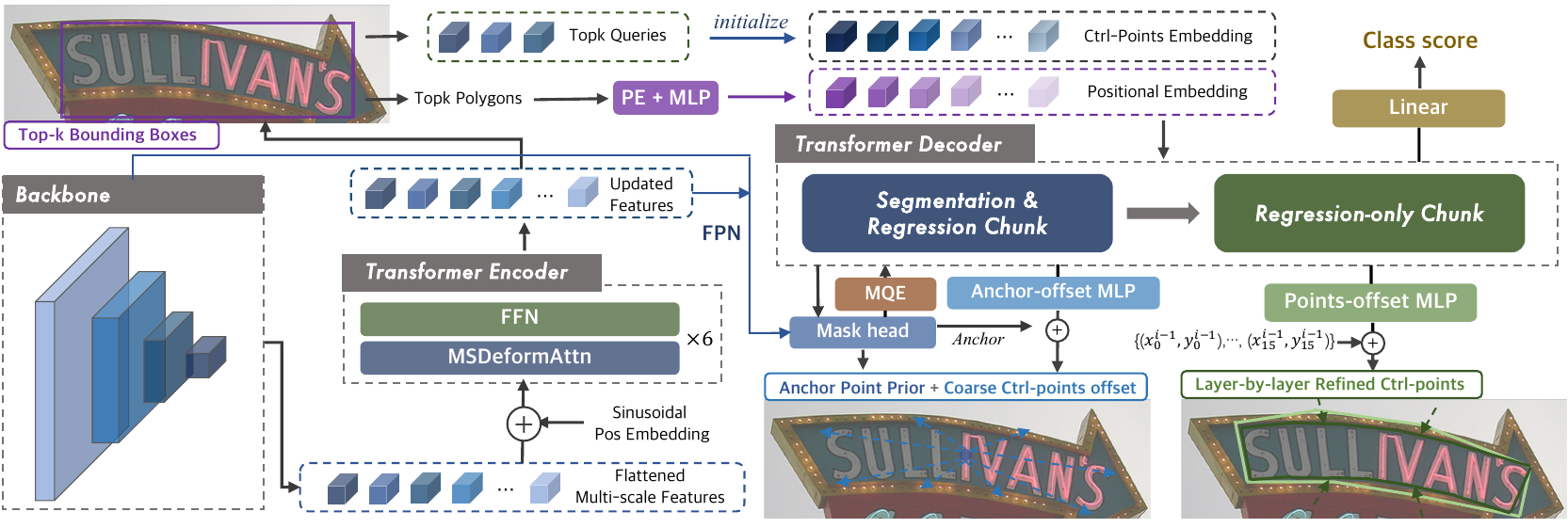}
    \caption{The overview of the proposed SRFormer. 
    % Initially, the flattened multi-scale backbone feature undergoes a progressive enhancement process through a 6-layer deformable transformer encoder. Subsequently, the enriched feature maps are integrated with the original ones in a top-down hierarchical manner with a FPN, to facilitate semantic segmentation and serve as a reference for subsequent decoding operations. The decoder embeddings are derived from the encoder queries and a set of learnable parameters, which are then processed through 
    We propose a two-step mechanism in the decoder: firstly, acquiring a coarse positional prior with segmentation results, followed by iterative regression refinements in a layer-by-layer fashion. We aim to optimize the intermediate representations and final predictions for improved performance in a concise framework.}
    \label{fig:overview}
\vspace{-0.5cm}
\end{figure*}

\section{2. Related Work}

\subsection{2.1. Detection Transformers}

DETR (DEtection TRansformer)~\cite{carion2020end} represents a pioneering model that introduced a fully end-to-end transformer-based paradigm for object detection. By formulating object detection as a set prediction task, it eliminates the need for non-maximum suppression (NMS) and substantially reduces post-processing requirements. However, DETR's training convergence and feature resolution limitations have hindered its competitiveness compared to traditional detectors. In response, Deformable DETR~\cite{zhu2020deformable} addresses these concerns by introducing sparse multi-scale features to enhance efficiency. Additionally, other variants such as Conditional-DETR~\cite{meng2021conditional}, Anchor-DETR~\cite{wang2022anchor}, and DAB-DETR~\cite{liu2022dab} introduced improved positional priors to expedite the training process. Furthermore, approaches like Group-DETR~\cite{chen2022group} and DN-DETR~\cite{li2022dn} concentrate on label assignment strategies, significantly improving matching stability, particularly during early training stages. Our study primarily focuses on the transformer decoder part, aiming to enhance the quality of query representation and expedite the training convergence.

\subsection{2.2. Segmentation-based Scene Text Detectors}

Segmentation-based approaches commonly integrate pixel-level prediction with subsequent post-processing algorithms to obtain the bounding boxes or polygons corresponding to the detected objects. CRAFT~\cite{baek2019character} utilizes a weakly supervised approach to train character segmentation models. PSENet~\cite{wang2019shape} first predicts the text center region (text kernel) and then obtains the result of text instance segmentation by progressive scale expansion algorithm. DBNet~\cite{liao2020real} embeds differentiable binarization into the network and predicts the corresponding threshold map in addition to learning the binary segmentation map of the text region. Learning low-level representations makes segmentation-based methods more robust towards domain gaps and font variations. However, the total inference time is considerably impacted by post-processing operations on the CPU. Our proposed model seamlessly integrates the prowess of representation learning, while being free from the need for intricate post-processing.

\subsection{2.3. Regression-based Scene Text Detectors}

Regression-based methods directly predict the polygon coordinates or Bezier control points. EAST~\cite{zhou2017east} represents an end-to-end anchor-free method that adopts pixel-level regression techniques to handle multi-oriented text instances. ABCNet~\cite{liu2021abcnet} is the first to introduce Bezier curve control points for arbitrary-shaped texts. TESTR~\cite{zhang2022text} and DPText~\cite{ye2023dptext} exploit the efficacy of the DETR architecture, wherein they utilize learnable queries as inputs and employ a straightforward MLP head to predict polygon coordinates. We preserve the procedural simplicity inherent in the regression-based methods, while enhancing performance and robustness through a judicious fusion of segmentation.

\section{3. Methodology}

\subsection{3.1. Overview}

\paragraph{Model Architecture.}

Fig.\ref{fig:overview} shows the overall structure of \model. We first leverage ResNet50~\cite{he2016deep} as the backbone. Upon updating the flattened features with the transformer encoder, we combine the backbone and updated features with a feature pyramid network module. The fused features are then channeled into both the decoding stage and the mask prediction head, thereby serving as the foundational reference feature within the framework. This interaction between query representations and high-resolution backbone features addresses the information bottleneck observed in the original DETR segmentation heads. Subsequently, we employ a two-stage approach, where a shared group of decoder embeddings is initialized by encoder output and fed into the decoder to gather richer features through cross-attention mechanism. We set the first few layers as the Segmentation\&Regression Chunk to make instance-wise mask prediction along with the point-wise coordinate prediction, followed by Regression-only chunk performing layer-by-layer refinement to get more precise polygon control points. Several heads for mask, coordinate and class score predictions, are adopted in a parallel manner.

\paragraph{Query Formulation.}
Derived from previous success, we initialize decoder queries with encoder outputs for better performance and faster training convergence. Instead of setting the number of learnable parameters as the number of proposals $K$. We only set 16~(i.e., number of polygon control points) groups of learnable embeddings to capture point-wise feature and universal control point correlation. They are then equipped with top-k encoder queries $q_{e}\in \mathbb{R}^{K\times d}$ to provide instance-wise information:
\begin{equation}
q_{d} = q_{e}[\texttt{Arg\_Top\_K}(s_{cls})] +q_{p}
\end{equation}
where $q_{d}\in \mathbb{R}^{K\times 16\times d}$ is the decoder embedding, $s_{cls}$ denotes the classification score predicted from encoder queries, and $q_{p} \in \mathbb{R}^{16\times d}$ is point-wise learnable embedding. By combining instance-level and control point-level queries to form a hierarchical representation, we can effectuate the filtration of similar predictions through instance-level attention, and model global point-to-point relative relationships through point-level attention. 

In addition, to better utilize the bounding box output from encoder, we sample 16 equidistant point uniformly along the longer side of the box in a clock-wise manner as proposed in~\cite{ye2023dptext}. These sampled points are subsequently employed as the initial polygon prediction.  We use sinusoidal positional encoding function $\texttt{PE}(\cdot)$ in conjunction with a two-layer MLP scaling network $\texttt{MLP}(\cdot)$ to enable precise positional representation for each control point:
\begin{equation}
   q_{pos} = \texttt{MLP}(\texttt{PE}(p_{xy})) \in \mathbb{R}^{K\times 16\times d}
\end{equation}
where $p_{xy}\in \mathbb{R}^{K\times 16\times 2}$ represents the coordinate of all polygon control points. 
\subsection{3.2. Segmentation \& Regression Chunk}

\begin{figure}[t]
    \centering
    \includegraphics[width=0.9\linewidth]{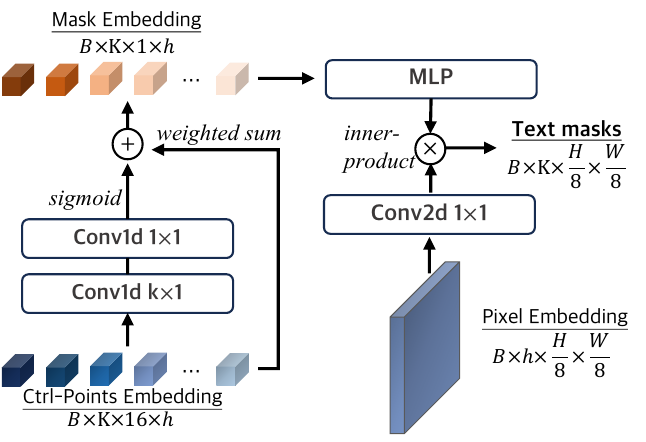}
    \caption{A detailed structure of mask prediction head.}
    \label{fig:mask_head}
    \vspace{-0.5cm}
\end{figure}

\paragraph{Mask prediction.}
As demonstrated in Fig.~\ref{fig:overview}, we only perform text instance segmentation at initial layers of decoder, based on the experimental findings that instance segmentation masks show favorable results in first few layers and can hardly be refined layer-by-layer even with improved query representations in deeper decoder layers. With this implementation, we can also reduce the computation cost in the decoder with minimal performance drop. To perform mask prediction, we build the pixel embedding map fused from backbone and encoder features. Given the hierarchical nature of the queries in the decoder, it becomes imperative to aggregate point-level queries for text instance-level prediction. We show a closer look of our mask head in Fig.~\ref{fig:mask_head}. Specifically, we first use a 1D Conv with large kernel sizes~($k=9$ in our default setting) to capture inter-point geometry knowledge, followed by a $1\times1$ Conv layer to learn point-level aggregation weights. Then we adopt weighted summation of queries along the control point dimension to adaptively formulate mask embedding. Finally, we dot-product each mask embedding $q_{m}$ with the pixel embedding map $ F^{1/8}$ to obtain instance masks $\hat{m}$:
\begin{equation}
\hat{m} = \mathcal{F}(q_{m}) \cdot \mathcal{P}(F^{1/8}) \in \mathbb{R}^{K\times H'\times W'}\ 
\end{equation}
where $\mathcal{F}$ denotes a two-layer MLP and $\mathcal{P}$ is a convolutional layer to make linear projection for semantic alignment.

\paragraph{Mask as regression prior.}
To bridge the gap of dense representation of segmentation masks and polygon control points, we first formulate dense anchor grids map $G \in \mathbb{R}^{H' \times W' \times 2}$ of the same resolution as segmentation masks:
\begin{equation}
\small
    \begin{aligned}
        G = \texttt{meshgrid}(&\texttt{linspace}(\frac{1}{H' + 1},\frac{H'}{H' + 1},H'),\\
                     &\texttt{linspace}(\frac{1}{W' + 1},\frac{W'}{W' + 1},W'))
    \end{aligned}
    % \nonumber
\end{equation}
where the $\texttt{linspace}(\textit{start, end, num)}$ function evenly generate \textit{num} points in the closed interval [\textit{start, end}]. Subsequently, we perform Hadamard product between anchor grids and normalized text segmentation results to obtain the `center of gravity' for each text instance:
\begin{equation}
\hat{p}_a = \sum^{H'W'}\text{softmax}_{H'W'}(\hat{m} / \tau) \odot G
\end{equation}
$\hat{p}_a\in \mathbb{R}^{B \times K \times 2}$ are anchor points for subsequent regression, and $\tau$ is a scaling factor set to 0.3. The mask results are normalized using the softmax function across the spatial dimension to ensure the output $\hat{p}$ falls within the interval of $[0, 1]$. Empirical analysis has demonstrated that text confined to a small spatial area exhibits anchors that are noticeably attracted to the central region of the image. A scaling factor is applied to enhance the discriminative contrast between the response values pertaining to text and non-text regions to mitigate the potential influence of non-textual areas, characterized by lower response values yet encompassing larger pixel extents, on the final anchor outcome. Given that most text instances are regular convex geometries, their center of gravity coincides with the geometric center, making it a suitable reference point for regression purposes.

% \if 1
\paragraph{Discussion.}
In addition, our approach differs from MaskDINO~\cite{li2023mask} for obtaining positional priors from instance masks. MaskDINO employs the boundary rectangles of connected regions within binary segmentation maps, which can prove inaccurate in text detection tasks, particularly during initial training stages. Instance segmentation masks are obtained by computing the representational similarity between query and pixel embeddings, potentially leading to multiple responses surpassing a given threshold. This limitation becomes more pronounced in text detection, primarily due to the abundance of visually akin instances, especially in scenarios featuring high text density, such as documents. 
Our methodology inherently does not necessitate highly precise mask predictions.
% \fi

% \subsection{3.3. Regression-only Chunk}

\subsection{3.3. Mask-informed Query Enhancement}

Previous work~\cite{liu2022dab} revealed that cross-attention mechanism in the decoder can be treated as soft ROI pooling where the ROI is implicitly encoded in the positional embedding. We heuristically adopt instance mask as a kind of soft ROI to extract instance-level features from encoder features and add the feature directly to the original query representation, working as a supplementary of point-level features extracted through cross-attention. Specifically, we build instance-level ROI indicator $\hat{r}\in \mathbb{R}^{K \times H' \times W'}$ from mask prediction of all scales by:
\begin{equation}
    \hat{r} = \text{softmax}_{K}(\hat{m}) \odot \widetilde{m}
\end{equation}
we introduce the semantic segmentation mask $\widetilde{m}$ to achieve two objectives: firstly, to softly filter out non-textual regions within $\hat{m}$, and secondly, to incorporate supplementary supervision as proposed in~\cite{long2022towards}. The ROI for each instance is softly excluded from each other, thereby augmenting the differentiation among query representations and facilitating model optimization. Subsequently, we extract instance-level and global text features from multi-scale encoder features normalized by the spatial-wise summation of ROI indicators to craft the final output $I\in \mathbb{R}^{K \times d}$ for query enhancement:
\begin{equation}
    I = \texttt{MHA}(\frac{\sum^{L}\sum^{H'W'}(\hat{r}^{l} \odot F^{l})}{\alpha \cdot L}) + \frac{\sum^{H'W'}(\mathbf{\Gamma}(\widetilde{m}) \odot F^{1})}{\beta}
\end{equation}
where $F^{l}$ represents encoder features of level $l$, the normalization factors $\alpha, \beta$ are formulated as $\sum^{H'W'}\hat{r}$ and $\sum^{H'W'}\mathbf{\Gamma}(\widetilde{m})$ respectively, $\mathbf{\Gamma}$ denotes a simple interpolation function to perform spatial alignment and $\texttt{MHA}(\cdot)$ represents a multi-head attention module to capture inter-instance relations. After applying linear projection to align with the original queries, we integrate the output of the MQE module directly into the query tensor. Enhanced queries are then directed into the subsequent decoder layers.

\begin{figure}[t]
    \centering
    \includegraphics[width=0.94\linewidth]{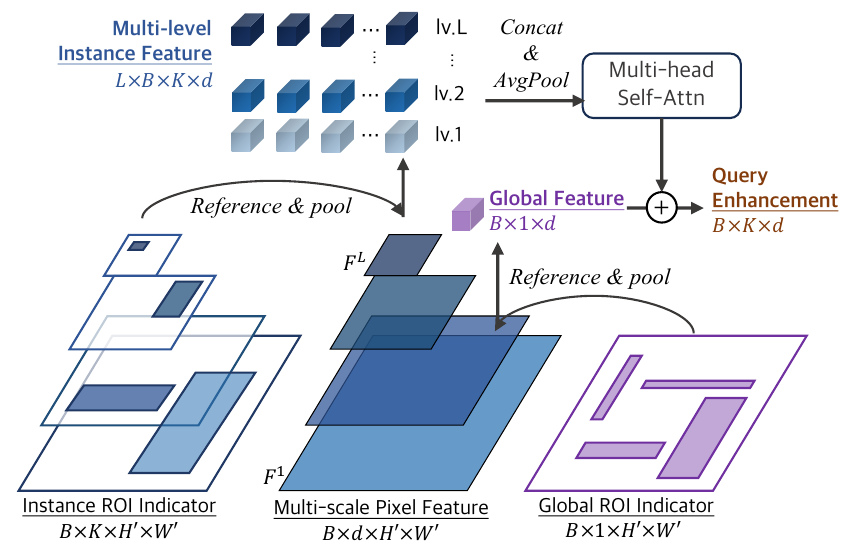}
    \caption{The Mask-informed Query Enhancement~(MQE) module extracts multi-level pixel features guided by instance and global ROI indicators, which are derived from instance and semantic segmentation masks respectively. }
    \label{fig:MQE}
\vspace{-0.5cm}
\end{figure}

\begin{table*}[t]
    \centering
    \small
    \begin{tabular}{l|l|ccc|ccc|ccc}
    \toprule
         \multirow{2}*{Methods} & \multirow{2}*{External Data} &  \multicolumn{3}{c|}{Total-Text} & \multicolumn{3}{c|}{CTW1500} & \multicolumn{3}{c}{ICDAR19-ArT}\\
         \cline{3-11}
         % \hline
         & &  P & R & F1 & P & R & F1 & P & R & F1\\
         \midrule
         \midrule
         \multicolumn{11}{c}{Segmentation-based Methods} \\
         \midrule
         TextSnake~\cite{long2018textsnake} & {\scriptsize Synth800K} & 82.7 &74.5 &78.4 & 67.9 & 85.3 & 75.6 & - & - & -\\
         PAN~\cite{wang2019efficient} &{\scriptsize Synth800K} & 89.3 & 81.0 & 85.0 & 86.4 & 81.2 & 83.7 & 61.1 & \textbf{79.4} & 69.1\\
         CRAFT~\cite{baek2019character} & {\scriptsize Synth800K+IC13} & 87.6 & 79.9 & 83.6 & 86.0 & 81.1 & 83.5 & 77.2 &68.9 &72.9\\
         DB~\cite{liao2020real} & {\scriptsize Synth800K} & 87.1 & 82.5 & 84.7 & 86.9 &80.2 &83.4 & - & - & -\\
         I3CL~\cite{du2022i3cl} & {\scriptsize  Synth800K(+MLT17+LSVT)} & 89.2 & 83.7 & 86.3 & 87.4 & 84.5 & 85.9 & 82.7& 71.3 &76.6\\
         \midrule
         \midrule
         \multicolumn{11}{c}{Regression-based Methods} \\
         \midrule
         ABCNet v2~\cite{liu2021abcnet} & {\scriptsize Synth150K+MLT17} & 90.2 &84.1 &87.0 & 85.6 & 83.8 & 84.7 & - & -& -\\
         FSG~\cite{tang2022few} & {\scriptsize Synth800K} & 90.7 & 85.7 & 88.1 & 88.1 &82.4 &85.2 & - & - & -\\
         TESTR(polygon)~\cite{zhang2022text} & {\scriptsize Synth150K+MLT17} & \textbf{93.4} & 81.4 & 86.9 & \textbf{92.0} &82.6 &87.1 & - & - & -\\
         SwinTextSpotter~\cite{huang2022swintextspotter}& {\scriptsize Synth150K+MLT17+IC13+IC15} & - & - & 88.0 & - & - & 88.0 & - & - & -\\
         TextBPN++~\cite{zhang2023arbitrary} &{\scriptsize Synth800K+MLT17} & 91.8 & 85.3 & 88.5 & 87.3 &83.8 &85.5 & 81.1 &71.1 &75.8\\
         DPText~\cite{ye2023dptext} & {\scriptsize Synth150K+MLT17(+LSVT)} & 91.3 & \underline{86.3} & 88.7 &91.7 & \underline{86.2} &88.8 & 83.0 &73.7 &78.1\\
         \midrule
         \rowcolor{mygray}Ours-SRFormer (\#1Seg) & {\scriptsize Synth150K+MLT17(+LSVT)} & 92.2 & 86.6 & 89.3 & 91.6 & 87.7 & \textbf{89.6} & \textbf{86.2} & 73.1 & 79.1\\
         \rowcolor{mygray}Ours-SRFormer (\#2Seg) & {\scriptsize Synth150K+MLT17(+LSVT)} & 92.2 & \textbf{87.9} & \textbf{90.0} & 89.4 & 89.6 & 89.5 & \textbf{86.2}& 73.4&\textbf{79.3}\\
         \rowcolor{mygray}Ours-SRFormer (\#3Seg) & {\scriptsize Synth150K+MLT17(+LSVT)} & 91.5 & \textbf{87.9} & 89.7 & 89.4 & \textbf{89.8} & \textbf{89.6} & 86.1 & 73.5 & \textbf{79.3}\\
    \bottomrule
         
    \end{tabular}
    \caption{Quantitative detection results on several benchmarks. “P”, “R” and “F1” denote Precision (\%), Recall (\%) and F1-score (\%), respectively. The backbone network is all ResNet50, except for SwinTextSpotter~(SwinT), PAN~(ResNet18), CRAFT and TextSnake~(VGG16). We use \#Seg to denote the number of decoder layers assigned to the Segmentation\&Regression chunk.}
    \label{tab:compare_sota}
\vspace{-0.5cm}
\end{table*} 

\subsection{3.4. End-to-End Optimization}
\paragraph{Matching.}
The primary objective of the matching process is to ascertain an optimal permutation $\sigma: [\hat{Y}] \xrightarrow{} [Y]$ of N elements that minimize the matching cost between the set predictions $\hat{Y}$ and ground truths $Y$:
\begin{equation}
    \mathop{\arg\min}\limits_{\sigma}\sum^{N}_{n=1}\mathcal{C}(\hat{Y}^{(\sigma(n))}, Y^{(n)})
\end{equation}
where $N$ is the number of ground truth instances per image.
We use Hungarian matching to solve the corresponding bipartite matching problem. Regarding the cost function design, we use GIOU loss and L1 loss for bounding boxes along with a variation of focal loss for classification scores within the context of encoder output:
% \vspace*{-1pt}
\begin{equation}
    \begin{aligned}
        C_{enc} =& \lambda_{cls}\text{FL}(\hat{s}^{\sigma(n)}_{enc}) + \lambda_{coord}\Vert \hat{b}^{\sigma(n)} - b_{gt}^{n} \Vert \\
        & + \lambda_{giou}\text{GIOU}(\hat{b}^{\sigma(n)}, b_{gt}^{n}) 
    \end{aligned}
    % \nonumber
\end{equation}
where $\lambda_{cls}$, $\lambda_{coord}$, and $\lambda_{giou}$ are hyper-parameters to balance different cost proportions, $\hat{b}^{\sigma(n)}=(x_{c},y_{c},w,h)$ is the bounding box prediction, $\hat{s}^{\sigma(n)}$ is the classification score for text instances and $FL$ is defined as the difference between the positive and negative term: $FL(x)=- \alpha (1-x)^{\gamma}log(x) + (1-\alpha)x^{\gamma} log(1-x)$. 
The cost function for decoder output is formulated as:
% \vspace*{-2pt}
\begin{equation}
    \begin{aligned}
        C_{dec} =& \lambda_{cls}\text{FL}(\hat{s}^{\sigma(n)}_{dec}) + \lambda_{coord} \Sigma^{16}_{i=1}\Vert \hat{p}^{\sigma(n)}_{i} - p_{i}^{n} \Vert \\
        &+ \lambda_{mask}\text{Dice}(\hat{m}^{\sigma(n)}, m_{gt}^{n})
    \end{aligned}
\end{equation}
where $\hat{m}^{\sigma(n)}$ is the mask prediction for text instances, $\hat{p}^{\sigma(n)}_{i}$ denotes coordination prediction of the $i$-th control point and $\lambda_{cls}$, $\lambda_{coord}$, and $\lambda_{giou}$ are balancing factors. 
% $FL$ is defined as the difference between the positive and negative term: $FL(x)=- \alpha (1-x)^{\gamma}log(x) + (1-\alpha)x^{\gamma} log(1-x)$. 
It is worth noting that Dice loss is exclusively incorporated within the layers of Segmentation \& Regression chunk.

% $$
% C_{enc} = \text{GIOU}(\hat{b}^{\sigma(n)}, b_{gt}^{n}) + \Vert \hat{b}^{\sigma(n)} - b_{gt}^{n} \Vert + \text{FL}(\hat{s}^{\sigma(n)}, s_{gt}^{n})
% $$
\paragraph{Loss function.}
We leverage the focal loss with $\alpha=0.25, \gamma=2$ for instance classification. Dice loss in corporation with BCE loss are exploited to supervise both instance and semantic mask predictions. In addition, L1 distance loss is used for regressed polygon control points:
\begin{equation}
    \begin{aligned}
        \mathcal{L} = &\lambda_{cls}\mathcal{L}_{cls}(\hat{s}^{\sigma(n)}_{dec}, s^n_{gt}) + \lambda_{mask}\mathcal{L}_{mask}(\hat{m}^{\sigma(n)}, m_{gt}^{n})\\
            &+\lambda_{mask}\mathcal{L}_{mask}(\widetilde{m}, \Sigma^{N} m_{gt}^{n}) +\lambda_{reg}\mathcal{L}_{reg}(\hat{p}^{\sigma(n)}, p^{n}_{gt}) 
    \end{aligned}
    % \nonumber
\end{equation}
where $\lambda_{cls}, \lambda_{mask}$ and $\lambda_{reg}$ are balancing factors.

\section{4. Experiment}

\paragraph{Datasets and benchmarks.}

\textbf{TotalText}~\cite{ch2017total} features a diverse range of text instances, including horizontal, multi-oriented, and curved text in natural scenes. The dataset contains over 1,500 high-resolution images with annotations, making it suitable for evaluating the robustness of text detection models across different text layouts and orientations. \textbf{Rot.Total-Text} constitutes a test set derived from the Total-Text test set, as initially proposed in~\cite{ye2023dptext}. We also integrate it to facilitate the development of optimal performance models.
\textbf{CTW1500}~\cite{liu2019curved} consists of 1,000 training images and 500 test images, with various text instances exhibiting diverse orientations, fonts, and perspectives. \textbf{ICDAR19-ArT}~\cite{chng2019icdar2019} is a large arbitrary-shape scene text benchmark, which includes multiple languages.
We also adopt the following additional datasets for pre-training:
\textbf{SynthText150k}~\cite{liu2020abcnet} is synthesized by overlaying computer-generated text on natural images. This approach allows for large-scale data generation and fine-grained control over text characteristics, such as size, font, and orientation. The dataset contains contains 94,723 images with multi- oriented texts and 54,327 images with curved texts, providing a rich resource for pre-training text detection models under various synthetic scenarios. \textbf{MLT17}~\cite{nayef2017icdar2017} is introduced as part of the ICDAR17 Robust Reading Competition, which is a multi-language large-scale scene text dataset. 

% \vspace{-0.5cm}
\paragraph{Implementation details.}

We adopt ResNet-50~\cite{he2016deep} as the backbone, followed by a deformable transformer encoder with 8 heads and 4 sampling points to update the features. We set the number of proposals to 100 and polygon control point embedding to 16. Model pre-training is made on a mixture of SynthText150K, MLT17 and TotalText dataset for a total number of 300k iterations. The starting learning rate is 1e-4 and decays to 1e-5 at the 240k iteration. We fine-tune our model on TotalText and CTW1500 with 30k iteration with learning rates set to 1e-4 and 5e-5 respectively, which is then divided by 10 at the 24k iteration. For evaluation on ICDAR19-ArT dataset, we also incorporate LSVT for pre-training, following~\cite{sun2019icdar}. 
AdamW with $\beta_1$ = 0.9, $\beta_2$ = 0.999 and weight decay of $10^{-4}$ is leveraged as the optimizer. 
The loss weights for classification, mask prediction and ctrl-points regression are set to $\lambda_{cls}$=2, $\lambda_{mask}=\lambda_{reg}=5$, respectively. 
Various data augmentation strategies, including random crop, random blur, brightness adjustment, and color alteration, are employed in the training process. 
We adopt multi-scale training strategy with the shortest edge ranging from 480 to 896, and the longest edge kept within 1,600, following most of previous studies. For evaluation, we resize the shorter side to 1,000 and keep the longer side within 1,800. All training and evaluation are conducted on a system with 8 NVIDIA 3090 graphics cards.

\subsection{4.1. Comparison with SoTA Methods}
Our proposed methodology is evaluated on three benchmark datasets, namely Total-Text, CTW1500, and ICDAR19 ArT. The obtained quantitative results are then systematically compared with those achieved by prior approaches, as illustrated in Table~\ref{tab:compare_sota}. Our method consistently achieves state-of-the-art performance across these datasets. We use \#Seg to denote the number of layers in the Segmentation \& Regression Chunk, where the total number of decoder layers stays 6. Compared to previous SoTA methods, for example, \model outperforms the state-of-the-art DPText by +1.3\%, +0.7\% and +1.2\% on TotalText, CTW1500 and IDCAR19-ArT respectively. Additionally, \model surpasses SoTA segmentation-based method I3CL by a notable gap of +2.7\%, +3.6\% and +2.7\% on three benchmarks.

\begin{table}[!t]
    \centering
    \small
    \setlength{\tabcolsep}{2.6mm}{
    \begin{tabular}{c c | c c c | c}
    \toprule

         \#Seg Layer & \#Reg Layer & P & R & F1 & FPS\\
         \midrule
         1 & 5 &88.6 &84.5 &86.5 & 9.7\\
         2 & 4 &89.0 &85.1 &87.0 & 8.6\\
         3 & 3 &88.0 &86.1 &87.1 & 7.9\\
         % 3 & 6 & & &\\
    \bottomrule

    \end{tabular}}
    \caption{Ablation results of several variations of \model with different decoder layers allocation.}
    \label{tab:segLayers}
\end{table}
% We've downloaded the officially provided weights of DPText and evaluated it on our system, which yields a result slight lower than its reported one~(88.7 v.s. 89.0) due to hardware issues.
\begin{table}[!t]
    \centering
    \small
    \begin{tabular}{c c | c c |  c | c }
    \toprule

         AnchorReg & MQE & F1 & Improv. & Extra Param. & FPS\\
         \midrule
          &  & 85.5 & - & - &10.5\\
         \checkmark &  & 86.0 & 0.5 & 0.39M & 10.5\\
          & \checkmark & 86.7 & 1.2 & 2.95M & 7.9\\
         \checkmark & \checkmark & 87.1 & 1.6 & 3.34M & 7.9\\
    \bottomrule
    \end{tabular}
    \caption{Ablations on test sets with \model(\#3Seg). ``AnchorReg" denotes the control point regression based on mask-generated anchor points. ``MQE" represents our proposed Mask-guided Query Enhancement module.}
    \label{tab:ablation}
\vspace{-0.5cm}
\end{table}

\subsection{4.2. Ablation Studies}
All the ablation experiments are conducted on TotalText without any pre-training. All models, unless specified otherwise, are trained for 50K iterations.

\paragraph{Decoder layer number.}
In this study, we undertook an investigation into the impact of varying the number of layers assigned to Segmentation \& Regression Chunk on the final performance of the model. In general, placing greater emphasis on segmentation learning tends to yield improved recall rates, albeit at the potential cost of reduced precision, which can be attributed to the absence of a fine-grained, layer-by-layer polygon refinement process in the Regression-only Chunk. We've also noticed that 
the decoder's first layer achieves favorable segmentation results that can hardly be further improved in subsequent layers, which could partially explain the marginal performance gain by simply adding more segmentation layers. It's worth noting that our method yields a competitive 87.1\% F1-score with only 50k iteration training solely on TotalText.

\begin{figure}[!t]
  \centering
    \begin{subfigure}{0.95\linewidth}
      \centering   
      \includegraphics[width=\linewidth]{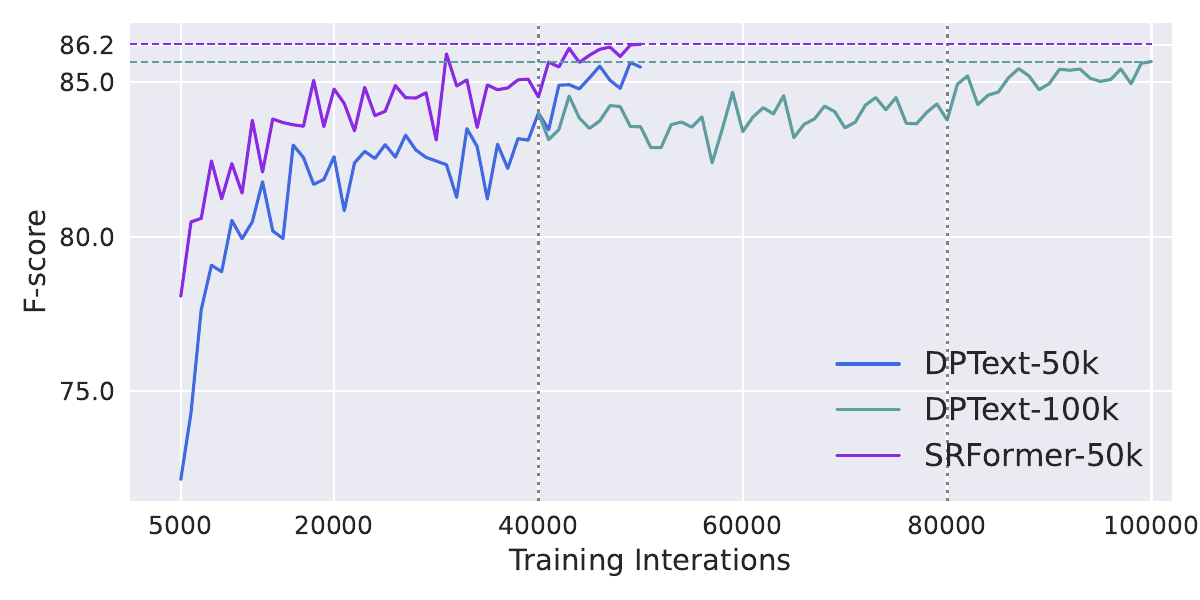}
        \caption{TotalText}
        \label{fig:sub1}
    \end{subfigure}   %      \hfill  % 这个\hfill指令为插入弹性长度的空白，看情况选择加不加。
    \begin{subfigure}{0.95\linewidth}
      \centering   
      \includegraphics[width=\linewidth]{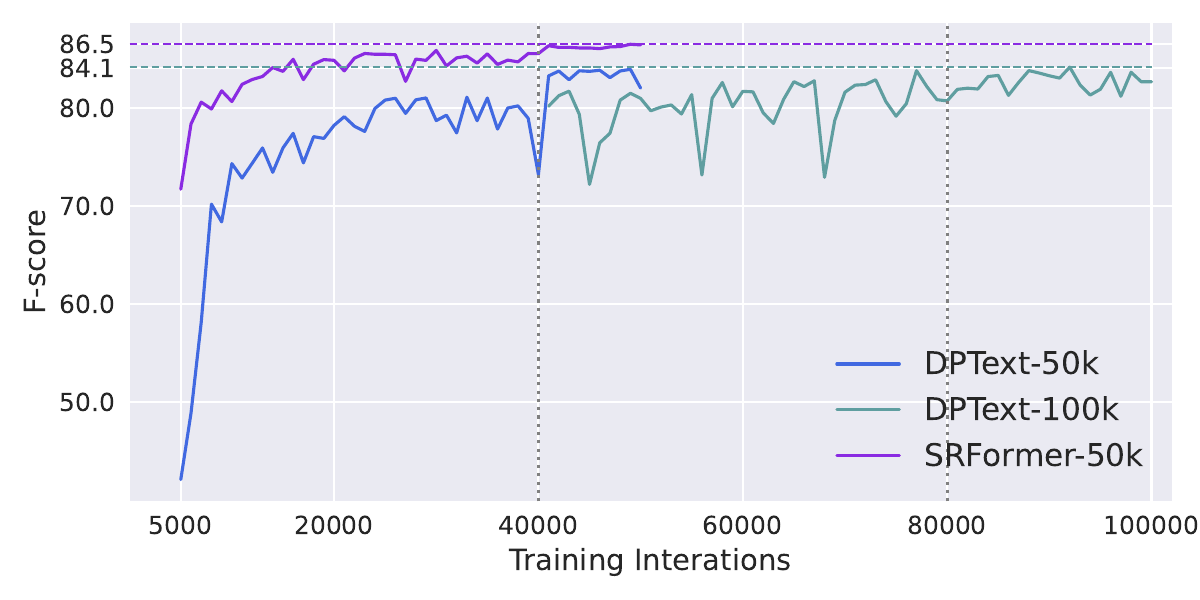}
        \caption{Rot.TotalText}
        \label{fig:sub2}
    \end{subfigure}
\caption{
Training convergence of DPText and ours. 
}
\vspace{-0.5cm}
\label{fig:training}
\end{figure}

% \paragraph{Unified matching.}

% \begin{table}[ht]
%     \centering
%     % \small
%     \begin{tabular}{c c | c }
%     \toprule

%          Dice & BCE & F1\\
%          \midrule
%           &  & \\
%          \checkmark &  &\\
%           & \checkmark & \\
%          \checkmark & \checkmark & \\
%     \bottomrule
%     \end{tabular}
%     \caption{Loss incorporated for unified matching.}
%     \label{tab:mathingloss}
% \end{table}

% Matching with both mask loss and reg loss or reg loss only

\paragraph{Regression from anchor points}

As listed in Table.~\ref{tab:ablation}, leveraging the anchor prior provided by instance masks brings about +0.5\% performance improvement. The utilization of mask-generated anchor points constitutes a valuable positional prior, especially at early training stages, enabling the model to learn geometric relationships and characteristics between control points.

\paragraph{Mask-informed Query Enhancement}

Incorporating MQE solely brings a notable +1.2\% performance gain, as shown in Table.~\ref{tab:ablation}
MQE module extracts distinctive pixel features for different queries by utilizing existing instance and semantic mask predictions, introducing less than 1M parameters at each layer. We believe that MQE can be treated as a cross-attention mechanism, where the mask functions analogously to positional embedding, guiding the model to extract richer features in a designated region.

\subsection{4.3. Discussion}
\if 0
Given that DPText represents the state-of-the-art in the domain of text detection, while also adopting the DETR-based model architecture and exhibiting a comparable number of parameters, it is employed as the exemplary regression-based approach for comparative analysis and discussion in this session. 
\fi 

\paragraph{Training efficiency.}

% \begin{figure}[t]
%     \centering
%     \includegraphics[width=\linewidth]{AuthorKit24/AnonymousSubmission/LaTeX/training_efficiency.pdf}
%     \caption{Training convergence curves evaluated on Total-Text for \model and previous SOTA method DPText. Two models are both trained from scratch.}
%     \label{fig:training}
% \end{figure}

%In order to illustrate the potential enhancement in training efficiency achievable through the incorporation of segmentation, we present convergence curves in Fig.~\ref{fig:training}, showcasing the fluctuation of the evaluation F1-score with increasing training iterations. 
Fig.~\ref{fig:training} shows convergence curves, showcasing the fluctuation of the evaluation F1-score with increasing training iterations.
When training from scratch on TotalText and Rot.TotalText, The observed trend in the figure reveals that our model consistently outperforms DPText in all tests beyond the 5,000th iteration, within the context of the 50k-iteration training schedule. In addition, we extended the training schedule of DPText twofold and generated its corresponding convergence plot. Despite the doubled training schedule, the performance of DPText still falls short of our model on both datasets. These findings emphasize the prospective advantages associated with the integration of segmentation, leading to enhanced convergence and superior performance in contrast to approaches solely reliant on regression methodologies.

\paragraph{Predictions at each decoder layer.}

\begin{table}[t]
    \centering
    \small
    \setlength{\tabcolsep}{2.7mm}{
    \begin{tabular}{c|cc|cc}
    \toprule
         test layer\# & Ours &  DPText & Ours$^{*}$ &  DPText$^{*}$\\
         \midrule
         layer 0& \textbf{86.67} & 81.64 & \textbf{82.13}$_{\text{(+26.96)}}$ & 55.17\\
         layer 1& \underline{\textbf{88.95}} & 86.67 & \textbf{84.32} & 76.87\\
         layer 2& \textbf{89.33} & 87.98 & \underline{\textbf{85.75}} & 83.86\\
         \midrule
         layer 5& \textbf{89.96} & 88.72 & \textbf{86.23} & 85.64\\
    \bottomrule
    \end{tabular}}
    \caption{F1 scores (\%) on TotalText when using different decoder layers in the same model. ${*}$ denotes models trained from scratch for 50k iterations.}
    \vspace{-0.2cm}
    \label{tab:perLayer}
\end{table}

\begin{table}[t]
    \centering
    \small
    \begin{tabular}{c|ccc|ccc}
    \toprule
         \multirow{2}*{Data Proportion}&  \multicolumn{3}{c|}{DPText} & \multicolumn{3}{c}{\model(Ours)}\\
         \cline{2-7}
         & P & R & F1 & P & R & F1 \\
         \midrule
         10\% Data& \textbf{83.0} & 69.3 & 75.6 & 82.9 & \textbf{71.8} & \textbf{76.9}\\
         50\% Data & 86.7 & 77.7 & 82.0 & \textbf{87.4} & \textbf{81.4} & \textbf{84.3}\\
    \bottomrule
    \end{tabular}
    \caption{Evaluation performance~(\%) of DPText and our model on TotalText dataset with fewer labeled images for training provided.}
    \label{tab:lowData}
\vspace{-0.5cm}
\end{table}

We take a closer look at the output of each decoder layer in \model and DPText to further reveal potential benefits brought by combined segmentation and regression, as listed in Table.~\ref{tab:perLayer}. In the context of pretrained models, our method produces SoTA result of 88.95\% F1-score with only two decoder layers, exceeding the six-layer DPText. For models trained from scratch, this gap is even more pronounced. The F1-scores of the predictions made by first decoder layer~(referred to as ``layer 0" in the table) exhibit a noteworthy gap of 26.96\%. 

\begin{figure}[t]
    \centering

        \includegraphics[width=0.95\linewidth]{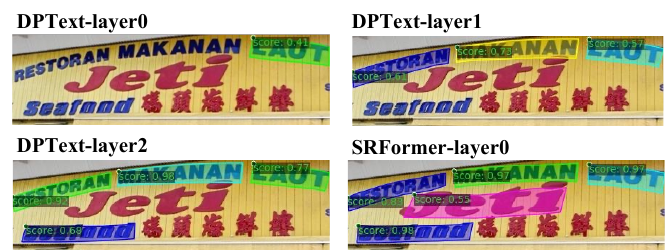}

    \caption{Per-layer visualization results of DPtext and \model on a test image from TotalText.}
    \label{fig:perlayer_vis}
    % \vspace{-0.5cm}
\end{figure}

% \if 0
We also conduct qualitative analysis to support our claim. As shown is Fig.~\ref{fig:perlayer_vis}, while our model detects all the English text in the image in the first layer, DPText, despite its capabilities, detects only a singular occurrence of text in the first layer and fails to detect the distorted text in the middle of the image in the third layer. Achieving robust prediction performance within the initial layers constitutes a cornerstone for our model's eventual attainment of SoTA outcomes. 
Proficient discrimination between textual and non-textual in early stages lies a strong groundwork. Subsequent focus is directed towards progressive refinement of positional predictions, as the model advances towards its optimal state.
% Learning to segment enables our model to better distinguish between text and non-text regions.

% \fi

\begin{figure}[!t]
    \centering
    \begin{subfigure}{0.3\linewidth}
      \centering   
      \includegraphics[width=\columnwidth]{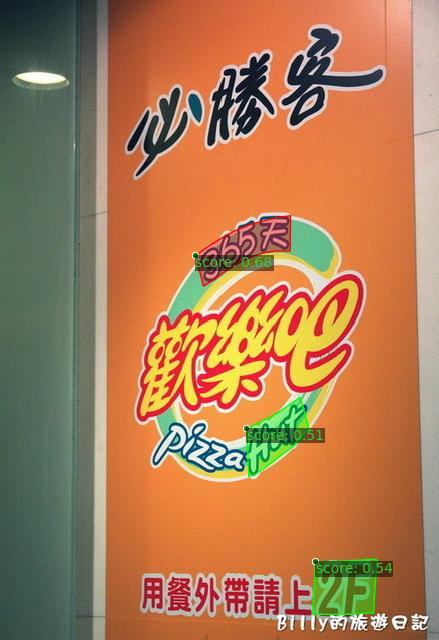}
        \caption{DPText}
        % \label{fig:sub1}
    \end{subfigure}
        \begin{subfigure}{0.3\linewidth}
      \centering   
      \includegraphics[width=\columnwidth]{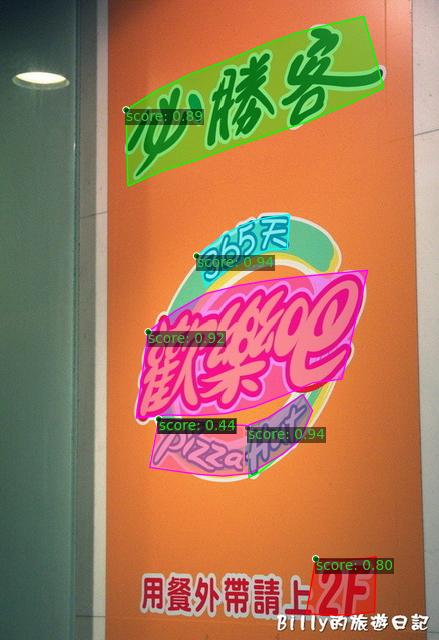}
        \caption{\model}
        % \label{fig:sub1}
    \end{subfigure}
        \begin{subfigure}{0.3\linewidth}
      \centering   
      \includegraphics[width=\columnwidth]{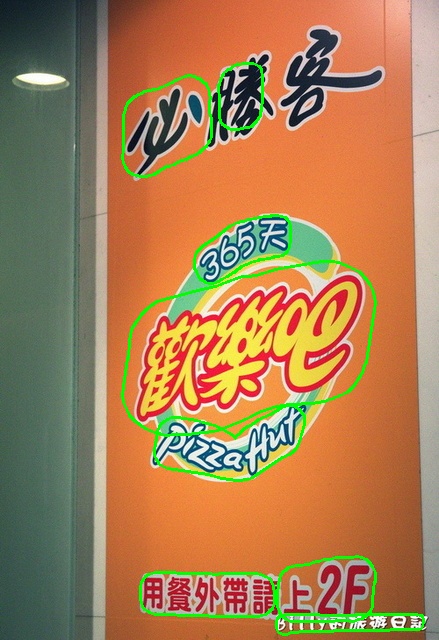}
        \caption{DBNet}
        % \label{fig:sub1}
    \end{subfigure}
    \caption{Visualization results for three models trained only on TotalText. The tested image is sampled from the test split of TotalText dataset.}
    \label{fig:compare}
    \vspace{-0.5cm}
\end{figure}

\begin{table}[t]
    \centering
    \small
    \begin{tabular}{l|cc|cc}
    \toprule
         \multirow{2}*{Training Set} & \multicolumn{2}{c|}{DPText} & \multicolumn{2}{c}{SRFormer (Ours)} \\
         % \cline{2-5}
         & MLT & TT & MLT & TT \\
         \midrule
         SynthText + TT + MLT & 70.54 & 80.93 & 71.11 & 81.81\\
         SynthText + TT & 50.10 & 67.40 & 63.28 & 71.96\\
         SynthText & 41.14 & 48.71 &42.87 & 51.52\\
         \bottomrule
    \end{tabular}
    \caption{Evaluation F1-score (\%) of DPText and \model on MLT17 and TotalText~(TT) dataset.}
    \label{tab:domain-robustnss}
% \vspace{-0.5cm}
\end{table}

\begin{figure}[th]
    \centering
    \includegraphics[width=0.9\linewidth]{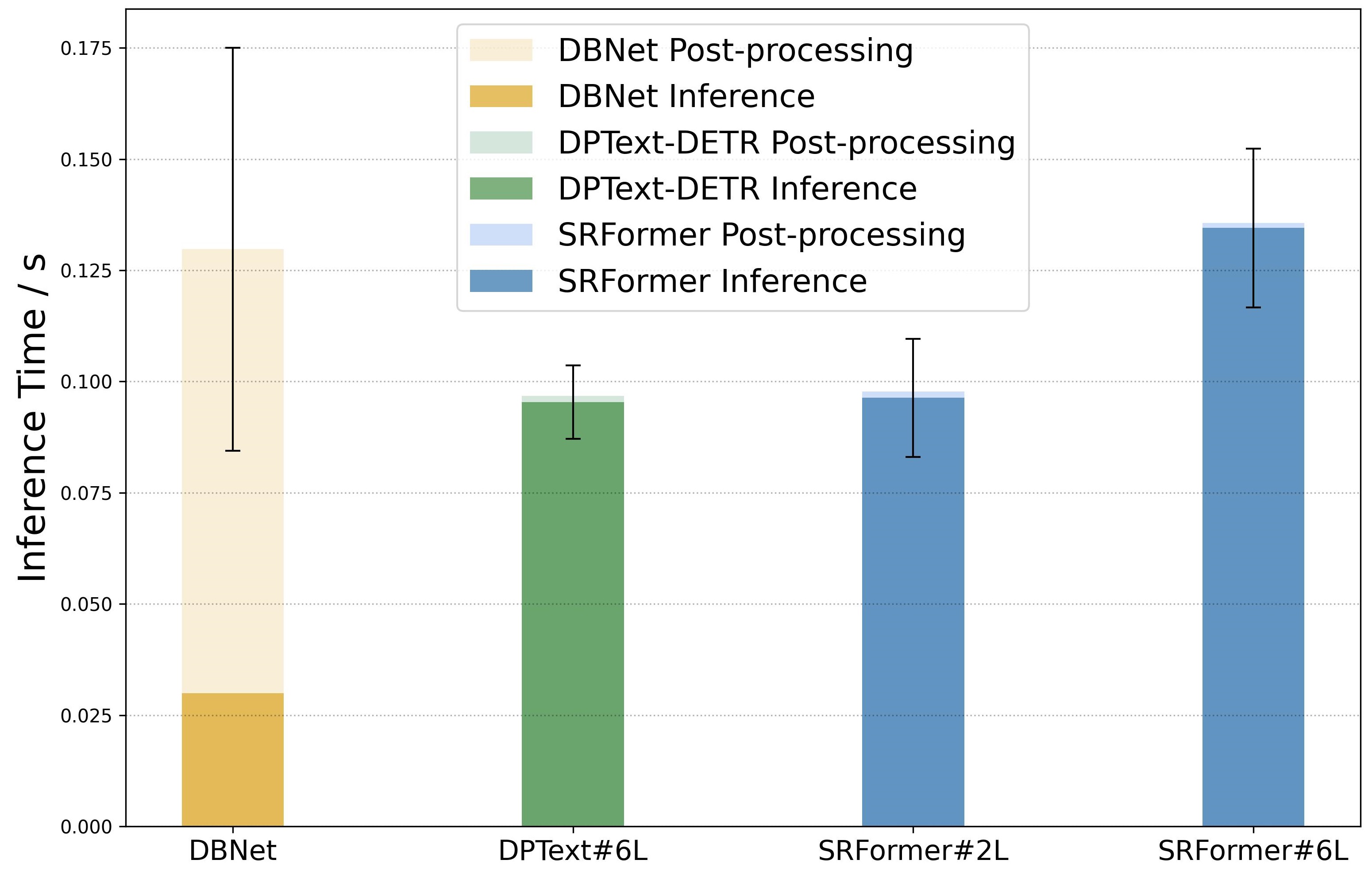}
    \caption{Average time spent on the GPU~(i.e., for inference) and CPU~(i.e., for post-processing) side per image. For SRFormer and DPText, we keep the longer side of input image within 1,800, while we resize the input to 1,600 for DBNet.}
    \label{fig:infer_time}
    \vspace{-0.5cm}
\end{figure}

\paragraph{Data efficiency.}
Segmentation-based techniques are commonly recognized for their superior data efficiency, demanding a considerably smaller training dataset to attain satisfactory generalization performance, which could be attributed to learning pixel-level representations by engaging with low-level features. Since our method utilizes a segmentation technique, our method can show better data efficiency. To reveal this property, we trained \model and DPText with only 10\% and 50\% samples in the TotalText dataset for 30k and 50k iterations, respectively. Table.~\ref{tab:lowData} shows the result. Using a limited training dataset comprising only 10\% of available samples, we achieved a F1-score of 76.9\%, demonstrating an improvement of approximately 1.9\% in comparison to DPText. Notably, upon increasing the training data availability to 50\%, the performance disparity further expands to 2.3\%, and our proposed method exhibits a F1-score of 84.3\%, underscoring its superior efficacy.

% The results reveal that our model demonstrates commendable generalization capabilities in low-data scenarios, with recall rates significantly surpassing those achieved by DPText~(e.g. +3.7 with 50\% data).

\paragraph{Robustness.}
As listed in Table.~\ref{tab:domain-robustnss}, we evaluate the cross-domain robustness inherent in two models by subjecting them to training and testing regimens involving disparate datasets. While TotalText addresses a more limited scope within scenes and languages, MLT, in contrast, encompasses a wide array of both domains. When trained on the combination of all, our model exhibits a relatively superior performance.
The exclusion of MLT data engenders an observable decrement in the performance of DPText. In contrast, our proposed model has an elevated level of robustness, evidenced by a significant performance upswing of +13.18\% on MLT and +4.56\% on TotalText, respectively. From an alternative vantage point, this performance differential tends to narrow when both models are only trained on synthetic data, reflecting SRFormer's capacity to cultivate real-world generalization even with limited samples. 

Furthermore, the qualitative analysis depicted in Figure \ref{fig:compare} demonstrates that our model, in conjunction with the segmentation-based approach DBNet, possesses a comparable ability for generalization in detecting languages beyond English when trained on a dataset primarily consisting of English content.

% \begin{table}[h]
%     \centering
%     \small
%     \begin{tabular}{c |c c c c c | c }
%     \toprule
%          Method & ARA & BAN & CH & JPN & KOR & EN-only \\
%     \midrule
%          \model & 45.0 & 72.3 & 35.2 & 37.5 & 50.5 & 62.8\\
%          DPText & 40.2 & 68.4 & 33.1 & 31.0 & 43.1 & 60.4\\
%     \midrule
%          diff. & 4.8 & 3.9 & 2.1 & \underline{6.5} & \textbf{7.4} & 2.4 \\ 
%     \bottomrule
%     \end{tabular}
%     \caption{Caption}
%     \label{tab:my_label}
% \end{table}

% \begin{table}[!h]
%     \centering
%     \small
%     \begin{tabular}{c|ccc|cc}
%     \toprule
%          & 1\#seg & 2\#seg & 3\#Seg & DPtext-50k & DPText-100k\\
%         \hline
%          F1&  63.3 & 63.6 & 63.1 & 59.6 & 59.4\\
%     \bottomrule
%     \end{tabular}
%     \caption{Evaluation results of TotalText-trained models on MLT17 dataset.}
%     \label{tab:robustness}
% \end{table}

% We show the superior robustness of model from the following three different perspective: generalize across different data sources, generalize across different languages, and low-data regime.

\paragraph{Inference time analysis.}
As mentioned in Sec~\ref{Sec:intro}, segmentation methods inherently require complex post-processing to obtain the final outputs from the identified segmentation map. While our method incorporates both segmentation and regression, the final output is determined by the regression head, eliminating the need for the post-processing step. Fig.~\ref{fig:infer_time} shows the required time for model inference and post-processing. The segmentation method, DBNet, incurs significant post-processing time, resulting in four times longer than model inference time and high variability per image. In contrast, the regression method, DPText, and ours demonstrate negligible post-processing time. Additionally, it's worth denoting that SRFormer\#2L, featuring only two decoding layers, exhibits a comparable inference cost to DPText but achieves better performances (as listed in Table.~\ref{tab:perLayer}).

% \vspace{-0.3em}

\section{5. Conclusion}
We propose SRFormer, a DETR-based model with incorporated segmentation and regression. By introducing mask prediction, we utilize it to provide a location prior for regression and to extract distinctive information for decoder queries from pixel features, enhancing robustness against textual deformations and improving domain transferability. Without compromising the simplicity of post-processing inherent to regression models, various experiments demonstrate that our method yields notable improvements in training efficiency, data utilization, and overall performance across various benchmarks. While the efficacy of the proposed method is substantiated within the context of text detection, we believe its prospective effectiveness in divergent detection tasks necessitating domain robustness and data efficiency.

\newpage

\section{Acknowledgement}
The work is supported in part by National Key R\&D Program of China (2022ZD0160201), HK RIF (R7030-22), HK ITF (GHP/169/20SZ), the Huawei Flagship Research Grants in 2021 and 2023, the HKU-SCF FinTech Academy R\&D Funding Schemes in 2021 and 2022, Hong Kong RGC GRF (HKU 17208223), and the Shanghai Artificial Intelligence Laboratory (Heming Cui is the Ph.D. advisor of Qingwen Bu and a courtesy researcher in this lab).

\bibliography{aaai24}

\end{document}